\documentclass[10pt,twocolumn,letterpaper]{article}

\usepackage{cvpr}
\usepackage{times}
\usepackage{epsfig}
\usepackage{graphicx}
\usepackage{amsmath}
\usepackage{amssymb}
\usepackage{subfigure}
\usepackage{multirow,tabularx}
\usepackage{authblk}

\usepackage[pagebackref=true,breaklinks=true,letterpaper=true,colorlinks,bookmarks=false]{hyperref}

\usepackage{amsthm}
\theoremstyle{plain}
\newtheorem{prop}{Proposition}
\theoremstyle{definition}

\cvprfinalcopy 

\def\cvprPaperID{3900} 

\ifcvprfinal\fi
\begin{document}

\title{Breaking the Spatio-Angular Trade-off for Light Field Super-Resolution via LSTM Modelling on Epipolar Plane Images}

\author[1]{Hao Zhu} 
\author[1]{Mantang Guo}
\author[2]{Hongdong Li}
\author[1]{Qing Wang}
\author[3]{Antonio Robles-Kelly}
\affil[1]{School of Computer Science, Northwestern Polytechnical University, China}
\affil[2]{Australian National University, Australia}
\affil[3]{Deakin University, Australia}

\maketitle

\begin{abstract}
Light-field cameras (LFC) have received increasing attention due to their wide-spread applications. However, current LFCs suffer from the well-known {\em spatio-angular trade-off}, which is considered as an inherent and fundamental limit for LFC designs. In this paper, by doing a detailed geometrical optical analysis of the sampling process in an LFC, we show that the effective sampling resolution is generally higher than the number of micro-lenses. This contribution makes it theoretically possible to break the resolution trade-off. Our second contribution is an epipolar plane image (EPI) based super-resolution method, which can super-resolve the spatial and angular dimensions simultaneously. We prove that the light field is a 2D series, thus, a specifically designed CNN-LSTM network is proposed to capture the continuity property of the EPI. Rather than leveraging semantic information, our network focuses on extracting geometric continuity in the EPI. This gives our method an improved generalization ability and makes it applicable to a wide range of previously unseen scenes. Experiments on both synthetic and real light fields demonstrate the improvements over state-of-the-art, especially in large disparity areas.

\end{abstract}

\section{Introduction}







The light-field camera\cite{lytro_web,raytrix_web} is becoming more and more popular. Due to its capability to capture whole 4D light field\cite{levoy1996light,gortler1996lumigraph} in a single shot, it enables new imaging capabilities such as refocusing\cite{ng2006digital} and free-viewpoint roaming. However, the performance of current LFCs is limited by the well-known \textit{spatio-angular trade-off}\cite{georgiev2006spatio}, namely, the notion that the product between the spatial resolution and angular resolution must not exceed the sensor resolution.


\begin{figure}[!t]
\begin{center}
\centering
	\includegraphics[width=80mm]{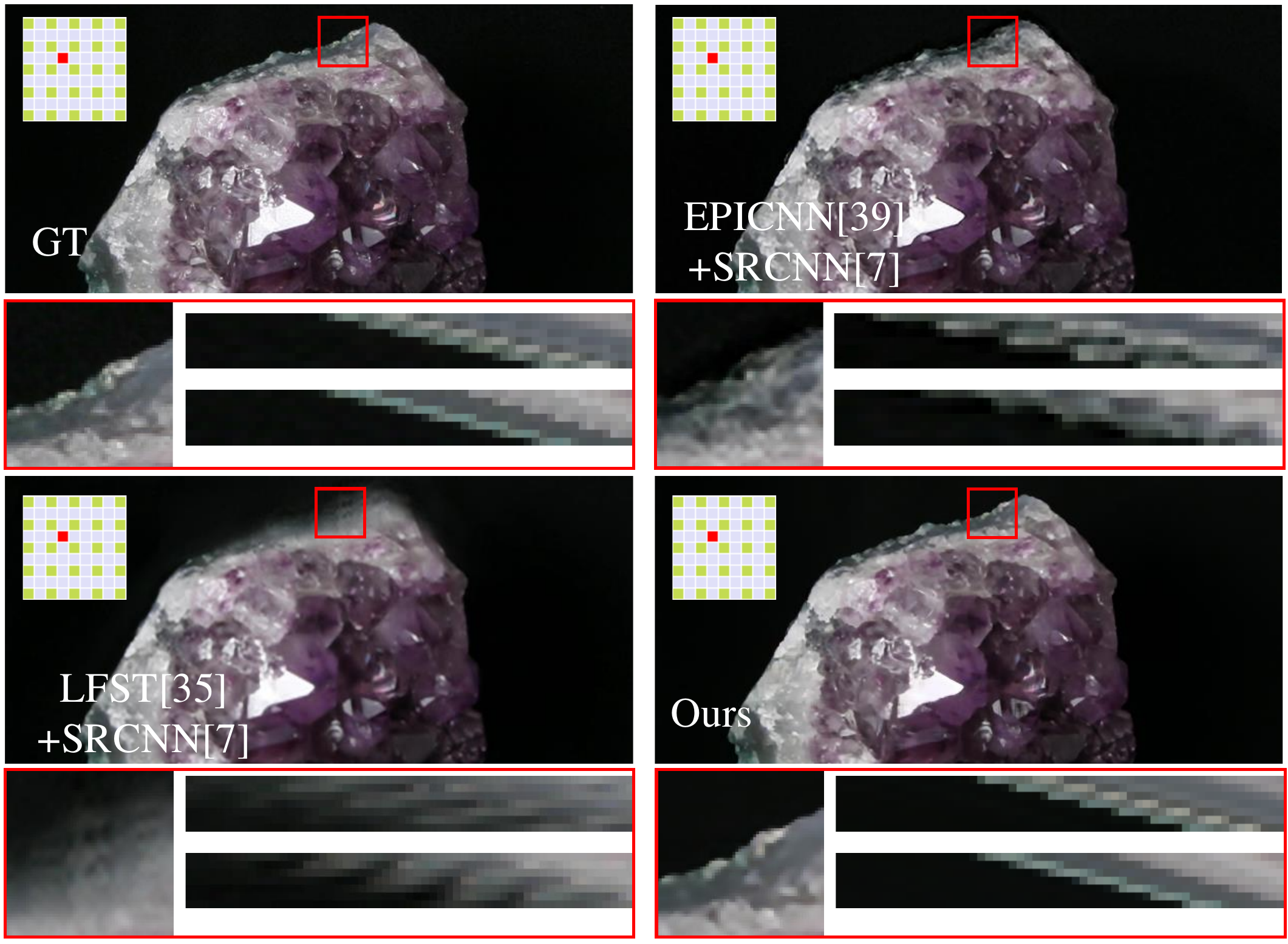}
\end{center}
\caption{Comparison of light field super-resolution of the Amethyst\cite{stanford_lf_web}. Given a low-resolution (sparsely sampled) input light field ($5\times5\times410\times307$), our method is able to produce a high resolution (densely sampled) light field ($9\times9\times820\times614$). The bigger picture in each sub-figure shows the reconstructed center view at $(4,4)$ as obtained a number of different methods. In the bottom row of each sub-figure, the left panel shows a close-up image region (as indicated by the red box in the full image). On the right panel, we show the reconstructed horizontal and  vertical EPIs.}
\label{fig:title_cmp}
\vspace{-0.5cm}
\end{figure}

To break the trade-off, several methods have been proposed to recover a high angular resolution light field from a low resolution input (Fig.\ref{fig:title_cmp}). However, there are still several challenges in current solutions. For depth-based methods\cite{eisemann2008floating,goesele2010ambient,chaurasia2011silhouette,chaurasia2013depth,penner2017soft,wanner2014variational}, the results are prone to errors in the depth estimation, which may cause artifacts on occlusion boundaries. Additionally, since each view is reconstructed independently, the geometric consistency between views can not be guaranteed. 

Recently, learning-based light field reconstruction has also been explored. Kalantari et al.\cite{LearningViewSynthesis} proposed two convolutional neural networks (CNNs)\cite{lecun1990handwritten,krizhevsky2012imagenet} to estimate depth and predict colors sequentially. However, since an explicit depth map has to be estimated, their method is still prone to estimation error. Wu et al.\cite{wu2017light} tackled the issues with depth based approaches by focusing on learning EPI super-resolution. They eliminated the information asymmetry\cite{yoon2015learning} between the spatial and angular dimensions by applying a blur operation on EPI. However, such a blur operation can not handle large disparity areas, where the continuous epipolar lines become discrete points. In this case, the information asymmetry still exists after the blur operation. Moreover, the EPI consistency is lost during the super-resolution process, which leads to fine structures in the image to be lost or over-smoothed in the reconstructed views.

In this paper, we first analyze the effective sampling resolution of an LFC and then propose a learning based method to super-resolve light fields in both, their angular and spatial dimensions. One of our key insights is that the {\em spatio-angular trade-off} only holds when the LFC is in {\em generalized focused case} (Sec.\ref{sec:LFD_model}). In the defocused case, 
the effective spatial sampling rate can be higher than the number of micro-lenses in the Plenoptic 1.0\cite{ng2006digital}. This insight is important since it provides the theoretical basis for further light field super-resolution beyond the resolution trade-off.

Another key insight is a learning-based framework for EPI super-resolution (Sec.\ref{sec:approach}). Although the appearance of epipolar lines varies (continuous vs discrete) in different disparities, they can be uniformly described with a 2D series model, which is the basis for introducing the well-known convolutional long short term memory (c-LSTM)\cite{hochreiter1997long,xingjian2015convolutional} for light field super-resolution. In contrast with previous super-resolution methods\cite{LearningViewSynthesis,wu2017light}, which leverage semantic information and content based inpainting, our network focuses on extracting and interpolating geometric continuity in EPI. This gives our method a better generalization ability and makes it applicable to a wide range of previously unseen scenes. Experiments (Sec.\ref{sec:experiments}) on both synthetic and real light fields demonstrate the performance of the proposed LSTM layers and hint at significant improvements over state-of-the-art learning-based methods ($>$3dB), especially in large disparity areas.

\section{Related Work}
\label{sec:related_works}
\noindent \textbf{Light field sampling:} Since the two-parallel-plane representation for light field sampling\cite{levoy1996light,gortler1996lumigraph} was proposed, two types of LFCs were developed, namely, the Plenoptic 1.0\cite{ng2006digital} and 2.0\cite{georgiev2009superresolution}. 
However, they both suffer the {\em spatio-angular trade-off}. Bishop et al.\cite{bishop2012light} analyzed the optical path in the Plenoptic 2.0. They pointed out that the aliasing effect in the spatial image contains new information, thus the resolution trade-off can be broken. The same conclusion was also summarized by Broxton et al. in \cite{broxton2013wave}, where the diffraction effects are proved to be helpful for improving lateral resolution of the light field microscope using wave optics. Compared with \cite{bishop2012light,broxton2013wave}, we focus on whole pixels instead of aliasing or diffraction, and prove that multiple views in the Plenoptic 1.0 record different point sets. The resolution of a light field can hence be improved by combining these point sets accordingly.

\noindent \textbf{Depth based methods:} Light field reconstruction can be viewed as a special case of image based rendering, as the input and reconstructed novel views are all restricted in a 2D grid. So previous depth based rendering techniques \cite{eisemann2008floating,goesele2010ambient,chaurasia2011silhouette,chaurasia2013depth,penner2017soft} can also be directly applied in light field reconstruction \cite{LearningViewSynthesis,srinivasan2017learning,wang2017light}. However, there are two problems in depth based algorithms. 
Firstly, there are depth ambiguities in shadow, reflection and refraction areas where a correct depth may not be a good depth. Secondly, as each view is reconstructed independently, the view consistency may be broken in the reconstructed light field. 


\noindent \textbf{Non-Depth based methods:} Considering the special grid features of light field sampling, some signal processing cues have been used in light field reconstruction. These include, but are not limited to the dimension gap between 3D focal stack and 4D light field \cite{levin2010linear}, the sparsity of light field sampling in continuous fourier domain\cite{shi2014light} and sparse representation of EPI in shearlet transform domain\cite{vagharshakyan2018light}.

Recently, CNNs have been used in light field reconstruction. Wu et al.\cite{wu2017light} tackled the light field reconstruction task viewing it as a one-dimensional EPI super-resolution problem and proposed the ``blur-restoration-deblur'' method. Wang et al.\cite{wang2018end} introduced the 4D CNN to directly super-resolve 4D light field instead of the 2D EPI. Yeung et al.\cite{yeung2018fast} explored the coarse characteristics of the sparsely-sampled light field and proposed the spatial-angular alternating convolutions to accelerate the reconstruction process. All of these CNNs treat light fields (or EPI) as a traditional 2D image, where each pixel is correlated with its standard ``4-neighboring system''. However, note the neighboring system size depends upon the direction of epipolar line in light field and its displacement. Thus, pixels with large disparities have a large neighboring system. As a result, previous CNN-based methods work well for narrow baseline light field while they often fail when applied to wide baseline light fields (see Fig.\ref{fig:title_cmp}).

\section{Optical Path Analysis in LFC}
\label{sec:LFD_model}
In this section, we prove that the well-known {\em spatio-angular trade-off} only exists when an LFC is in a {\em generalized focus case}, \ie, the disparities of all pixels in the recorded light field are integer values. In this case, all views in an LFC capture the same point set. Otherwise, different views account for different point sets which are aliased with respect to other views. As a result, 
the effective spatial resolution of the Plenoptic 1.0 is larger than the number of micro-lenses.

\begin{figure}[t]
\begin{center}
\centering
\subfigure[Generalized Focus case]{
	\label{fig:lfcamera_model:0}
	\includegraphics[width=74mm]{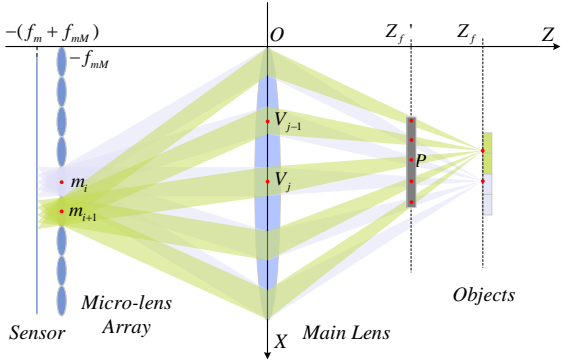} 
}\\
\subfigure[Defocused case]{
	\label{fig:lfcamera_model:1}
	\includegraphics[width=74mm]{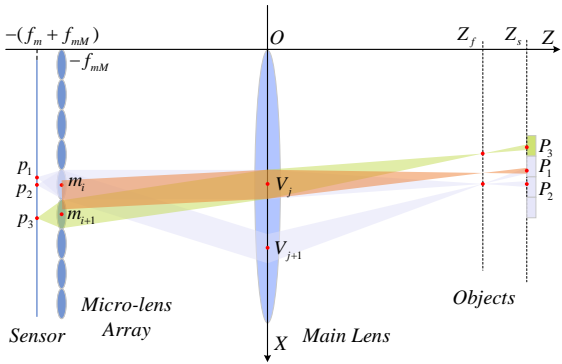} 
}
\end{center}
\caption{Optical Path in the Plenoptic 1.0.}
\label{fig:camera_model}
\end{figure}

\subsection{On the Number of Recorded Scene Points}
\label{sec:optical_path}
\noindent \textbf{Generalized focused case:}
In Fig.\ref{fig:lfcamera_model:0} we show the optical path of an ideal Plenoptic 1.0 camera, where all the pixels are covered by a micro-lens recording different views of a same point in space. In such case, the depth $Z_f$ of the scene point and the distance $f_{mM}$ between the micro-lens array (MLA) and the main lens must meet the Gaussian imaging principle, \textit{i.e.}, $\frac{1}{f_M}=\frac{1}{f_{mM}}+\frac{1}{Z_f}$, where $f_M$ is the focal length of the main lens. Here, the {\em spatio-angular trade-off} holds and all recorded pixels are clear images of objects at depth $Z_f$. That is, the recorded light field describes a consistent point set observed from each of the different views.

If the scene depth varies, \textit{i.e.}, the LFC is in the defocused case, where the pixels covered by a micro-lens become a uniform sampling over a circular area in space (the gray areas in Fig.\ref{fig:lfcamera_model:0}). In this case, different pixels under a micro-lens record different points in space. Note, however, that the above trade-off also holds in some defocused situations. When the point in space is moved to the depth $Z_f'$ in Fig.\ref{fig:lfcamera_model:0}, the point $P$ is still only recorded once by the micro-lens $m_{i+1}$ from view $V_j$. Other views also record it at different positions, \textit{e.g.}, view $V_{j-1}$ records it at the micro-lens $m_i$. In such a case, the images of point $P$ from different angles are also recorded at different micro-lenses (boundary pixels are ignored here). In other words, the recorded light field is still a multi-view description of a same point set.

The above defocused case is similar to the focused case in the sense that different views in the recorded light field describe the same set of scene points. We call both, the defocused and focused cases as {\em generalized focus case}, of which the mathematical formulation is given as,
\begin{equation}
    \forall p,\:d(p)\in \mathbb{Z},
\end{equation}
where $d(p)$ is the disparity of pixel $p$, $\mathbb{Z}$ is the set of integers. If all pixels in an LFC have integer disparities, the LFC is in {\em generalized focus case} and previous {\em spatio-angular trade-off} holds.

\noindent \textbf{Defocused case:}
Except for the above {\em generalized focus case}, different views of the recorded light field generally depict different scene point sets. As a result, the actual number of captured scene points is larger than the number of micro-lens.  Roughly speaking, the ``resolution-trade-off'' is broken in this case. In  Fig.\ref{fig:lfcamera_model:1}, we illustrate the defocused case. Pixels $p_1$ and $p_2$ under the micro-lens $m_i$ record two different points $P_1$ and $P_2$ from views $V_{j+1}$ and $V_{j}$, respectively. Note that the ray passing through the point $P_1$ from view $V_j$ to the MLA (the orange areas in Fig.\ref{fig:lfcamera_model:1}), is ``aliased'' by the micro-lenses $m_i$ and $m_{i+1}$. We can also trace the ray for the micro-lens $m_{i+1}$ from view $V_j$ to space point $P_3$. It can be seen that $P_1$ drops between points $P_2$ and $P_3$. Because $P_2$ and $P_3$ are the nearest points in view $V_j$, the point $P_1$, which is recorded by view $V_{j+1}$, is not recorded by view $V_{j}$. Thus, the number of effectively recorded points in LFC is larger than the resolution of view $V_j$. Since the resolution of each view equals the number of micro-lenses in the Plenoptic 1.0, the effective sampling resolution becomes larger than the number of micro-lenses.

\begin{figure}[!t]
\begin{center}
\centering
	\includegraphics[width=75mm]{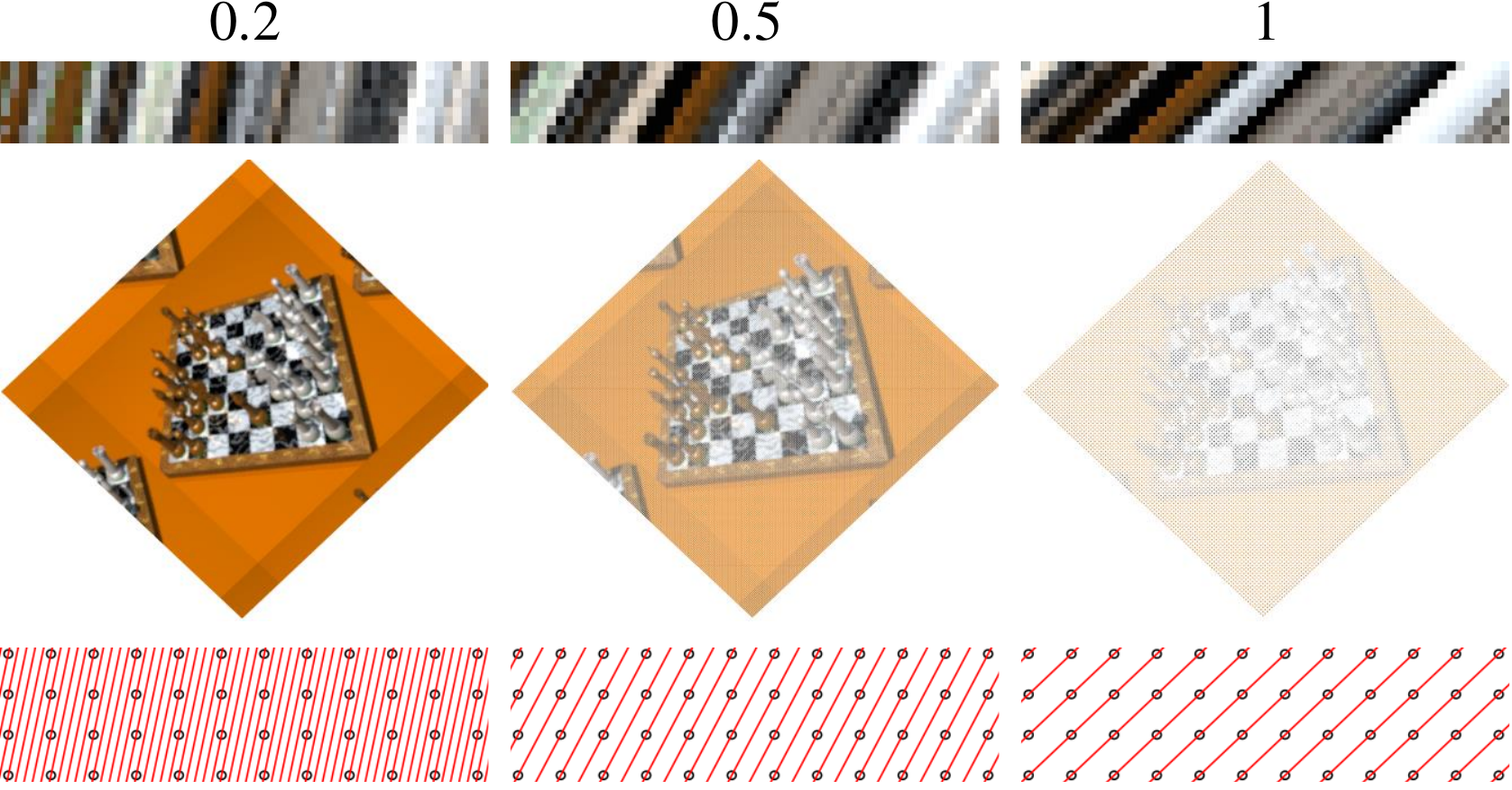}
\end{center}
\caption{The number of recorded pixels changes when the baseline in light field sampling also changes. From top-to-bottom we show the EPIs, the reconstructed point clouds and the sketch maps of light field sampling. From left-to-right we show the light fields with $0.2$, $0.5$ and $1$ pixel disparity. Note that the light field with $0.2$ pixel disparity records the larger number of points.}
\label{fig:camera_model_epi}
\end{figure}

We also provide an intuitive explanation to the above analysis on EPI. Fig.\ref{fig:camera_model_epi} shows EPIs and the corresponding point clouds under different disparity levels. Three light-fields are captured with different baselines. It is noticed that the number of recorded points are different in these light fields and the one with $0.2$ disparity records the most points. The sketches in the third row of Fig.\ref{fig:camera_model_epi} reveals the reason well. When the disparity is $0.2$, it can be seen that the red line passes through an entire pixel once every 5 views; in other words, views $\{1,5,9,...\}$ sample same point set while views $\{2,3,4,6,...\}$ sample other point sets. When the disparity equals to $1$, all views sample the same set of points. Thus, the light field with $0.2$ disparity records the largest number of scene points. To summarize:
\begin{prop}
\label{prop_res_trade_conclusion}
The {\em Spatio-angular trade-off} in LFCs only holds when the LFC is in the {\em generalized focused case}. This is due to the fact that the depth has a continuous and complex distribution in a real-world scene. Thus, the effective spatial resolution of the Plenoptic 1.0 is larger than the number of micro-lenses. In such cases, the light-field can be super-resolved.
\end{prop}

\section{Approach}
\label{sec:approach}
While the above sections show it is possible to recover high resolution light field defying the conventional spatio-angular trade-off, this is still a not straightforward task.   
The main difficulty strives from how to super-resolve a light field while keeping the consistency across different views. Most existing light field super-resolution methods are either based on depth recovery  \cite{zhou2018stereo,LearningViewSynthesis,srinivasan2017learning} or based on EPI analysis \cite{wu2017light,wang2018end,yeung2018fast}.  
The former approached are overly sensitive to errors in depth estimation, often failing to maintain cross-view consistency. The latter one treat EPIs as a regular digital image, failing to capture the EPI nature of continuous traces corresponding to pixels across multiple views.

In this section, we first discuss the issue of continuity preservation in light field super-resolution. We prove these continuities can be uniformly described as a 2D series. We then propose a novel CNN-LSTM architecture tailored for EPI super-resolution. 

\begin{figure}[t]
\begin{center}
\centering
	\includegraphics[width=55mm]{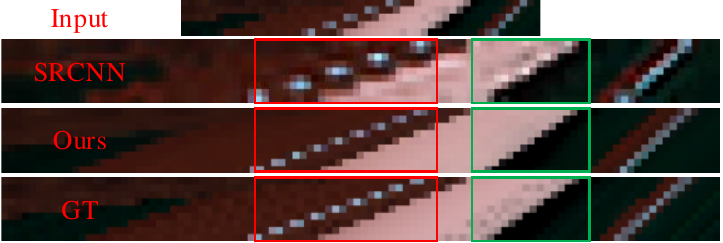}
\end{center}
\caption{Different types of continuity. The first, second, third and fourth rows show the input low resolution EPI, super-resolved EPI from \cite{dong2016image}, ours and ground truth, respectively. Previous image super-resolution CNN cope well with ``continuous continuity'' (green boxes), failing for ``discontinuous continuity'' (red boxes).}
\label{fig:dif_continuity}
\end{figure}

\subsection{Different continuities}
There are two types of continuities in light field, \textit{i.e.}, the `continuous continuity' and `jumping continuity' (Fig.\ref{fig:dif_continuity}). When the disparity is small, epipolar lines are continuous (green boxes), previous single image super-resolution methods\cite{dong2016image,kim2016accurate,lai2017deep} can be applied directly in such case. However, the continuous epipolar lines become discrete points (red boxes) when the disparity increases. Previous image super-resolution methods treat the discrete epipolar line as independent points, so epipolar line may be lost or over-smoothed in the super-resolved new views, leading thin structure objects missing or becoming unclear. Compared with `continuous continuity', the `jumping continuity' is more common in light field reconstruction task. Because novel views can be synthesized directly by interpolation in 4D space\cite{levoy1996light,chai2000plenoptic,lin2004geometric} when the disparity is small, it is unnecessary to use expensive reconstruction techniques.


\begin{figure}[h]
\begin{center}
\centering
\includegraphics[width=68mm]{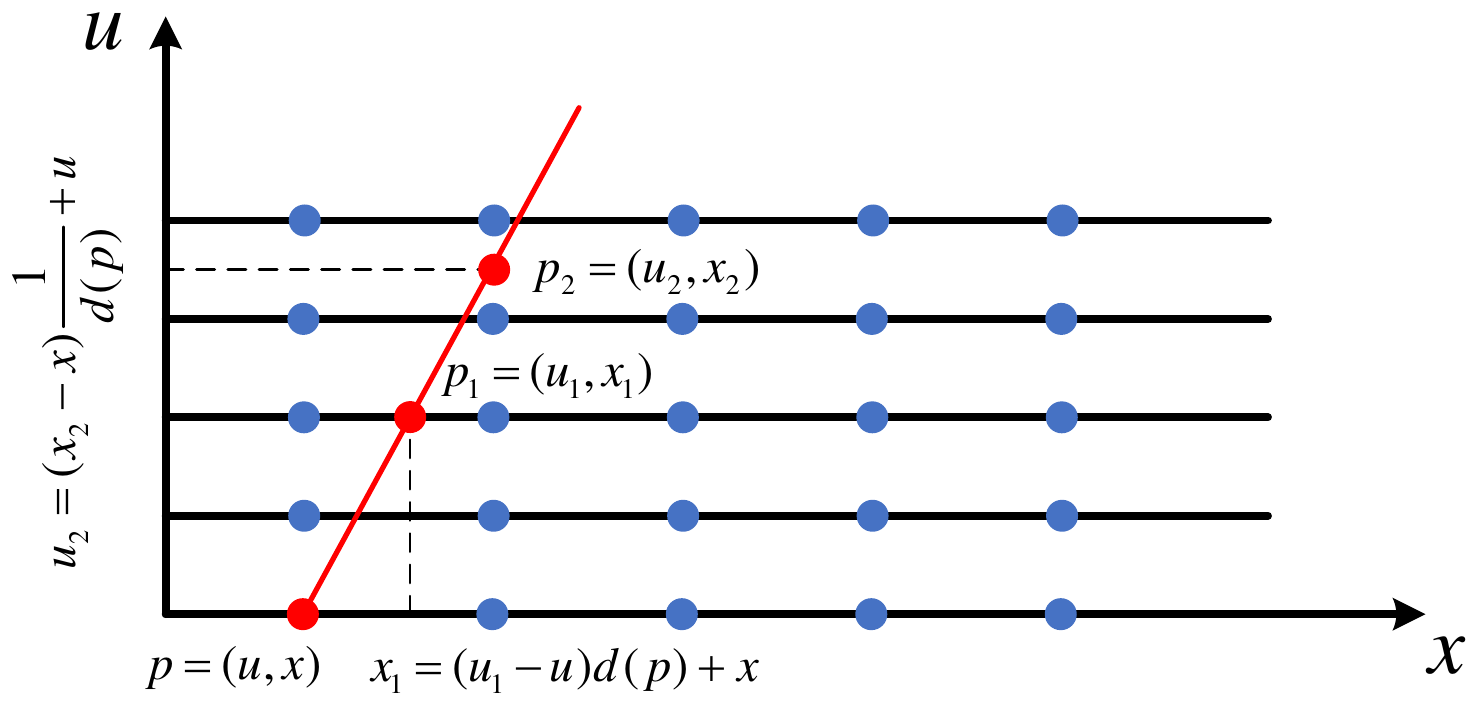}
\end{center}
\caption{For each epipolar line, it can be projected to angular or spatial axis when fixing on of the axes for $d\in(0,\infty)$.}
\label{fig:epi_time_series}
\end{figure}

\subsection{Light field as 2D series}
\label{sec:lf_timeseries}
Although the above two continuity cases appear to be distinct, they share a common characteristics, \ie, the pixels in these two different epipolar lines are all connected together through their disparities.

For each light ray $p=(u,x)$ in free space (without occlusion), there is a corresponding ray $p_1=(u_1,x_1)$ describing the same 3D point in any other view $u_1$, such that
\begin{equation}
x_1=(u_1-u)d(p)+x
\label{eqn:time_series_lambertian_1}
\end{equation}
where $d(p)$ refers to the disparity of $p$. In such case, the light field is a series in an angular space.

From another point of view, there is a corresponding ray $p_2=(u_2,x_2)$ in any other pixel position $x_2$, such that
\begin{equation}
u_2=(x_2-x)\frac{1}{d(p)}+u.
\label{eqn:time_series_lambertian_2}
\end{equation}
Here, the light field is also a series, but in a Cartesian space.

Fig.\ref{fig:epi_time_series} illustrates the above assertions, where each epipolar line can be projected to an angular or spatial axis by fixing one of the axes given $d\in(0,\infty)$. In other words, the ray in the light field is predictable if the disparity is known. Hence a light field may be treated as a 2D series allowing for the use of LSTM.

\begin{figure*}[!t]
\begin{center}
\centering
\includegraphics[width=165mm]{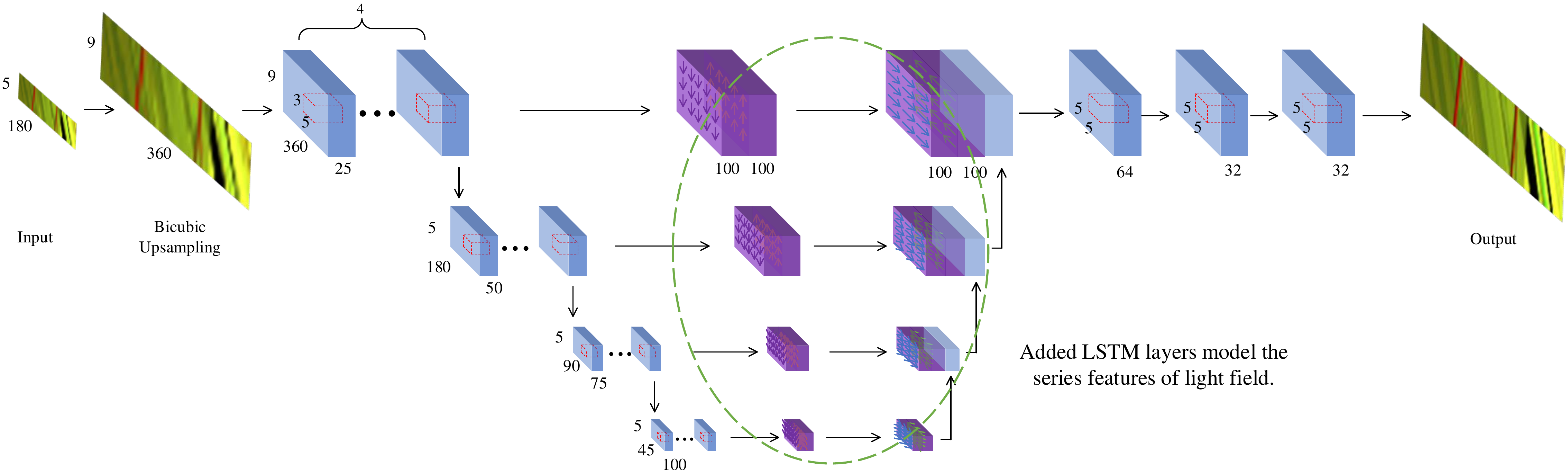}
\end{center}
\caption{The architecture of our neural network.}
\label{fig:cnnlstm_architecture}
\end{figure*}

\subsection{CNN-LSTM for EPI super-resolution}
Considering large disparities in the light field, we propose a CNN-LSTM network whose architecture is shown in Fig.\ref{fig:cnnlstm_architecture}. The overall network is inspired from the U-network in EPI analysis\cite{heber2016u,heber2017neural}. Our network has four ``levels'', where each of these accounts for the EPI at different resolutions. In contrast with previous work, four c-LSTM\cite{xingjian2015convolutional} layers are added at each level (the purple blocks in Fig.\ref{fig:cnnlstm_architecture}), to model the series nature of the EPI in the top-down, bottom-up, left-right and right-left directions, respectively. 
In our network, each c-LSTM has 100 channels while the kernel size is $1\times 3$.

When processing a low-resolution input EPI, this is firstly scaled 2 times up in both angular and spatial dimensions using Bicubic interpolation. Before LSTM analysis at each level, 4 convolutional layers are applied. These layers have kernels of size $3\times 5$. The channels of these convolutional kernels equal to $25\times i$ for the $i$-th level. After LSTM analysis, three convolution layers are added with kernel size $5\times5$ and channels $64$, $32$ and $32$. Note that each convolutional layer is followed by a ReLU \cite{nair2010rectified}. Different levels in Fig.\ref{fig:cnnlstm_architecture} are connected by down and up-convolutional layers with kernel size $3\times 3$.

It's worth noting in passing that the proposed light field super-resolution network takes a 2D EPI as input. Despite effective, this induces an inherent ambiguity as related to spatial super-resolution. The reason being that each EPI only covers one spatial dimension whereas every pixel is related to two spatial dimensions in the super-resolution process ($1$ pixel to $2\times 2$ pixels). This, however, as we will see later on in Section \ref{sec:experiments}, does not overly affect the performance of our method in practice.



\subsection{Datasets}
\label{sec:experiment:syn_data}
\begin{figure}[t]
\begin{center}
\centering
\includegraphics[width=75mm]{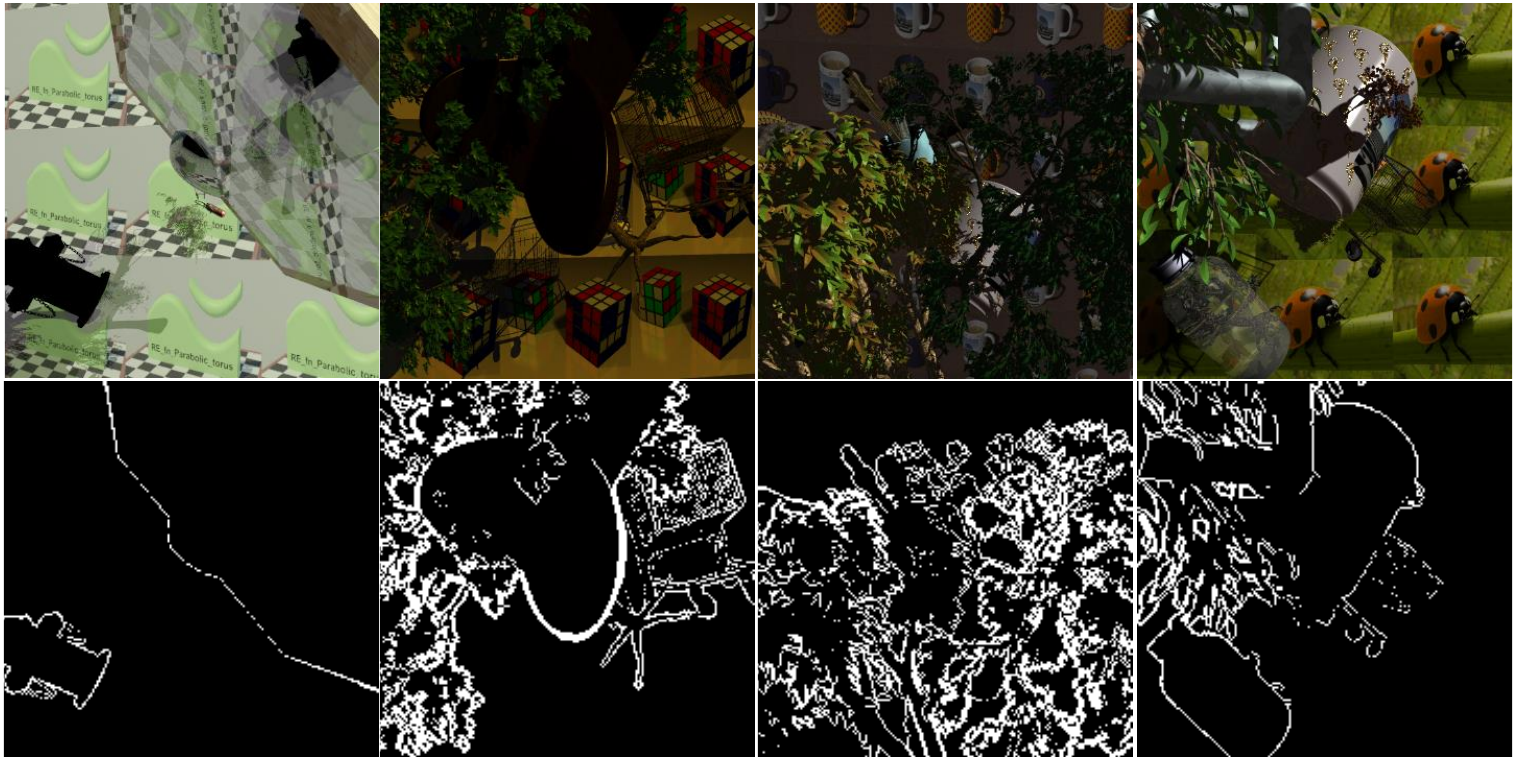} 
\end{center}
\caption{Light field examples. Top row: central views; Bottom row: corresponding occlusion maps.}
\label{fig:mydata}
\end{figure}

In order to train and evaluate our network, we build an automatic light field generator based on POV-ray\cite{povray_web,povray_resource} 
to render $100$ light fields. Fig.\ref{fig:mydata} shows some examples. For training and testing we have included various challenging environments in our dataset. These include inter-reflection, occlusion, shadowing, various illumination conditions and structures with fine detail. 

We augment the training data using two approaches. The first of these is exchanging the RGB channels. The second consists of shearing EPIs\cite{ng2006digital} using the expression
\begin{equation}
{EPI}_{d}(u,x)=EPI_{0}(u,x+ud),
\label{eqn:data_augmentation:shearing}
\end{equation}
where $EPI_{0}$ and $EPI_{d}$ are the original and sheared EPIs, respectively. The main goal of the shearing operation is to enhance the performance in negative disparity areas.

\begin{figure}[t]
\begin{center}
\centering
\subfigure[Recorded light field]{
	\label{fig:wrong_occ:a}
	\includegraphics[width=25mm]{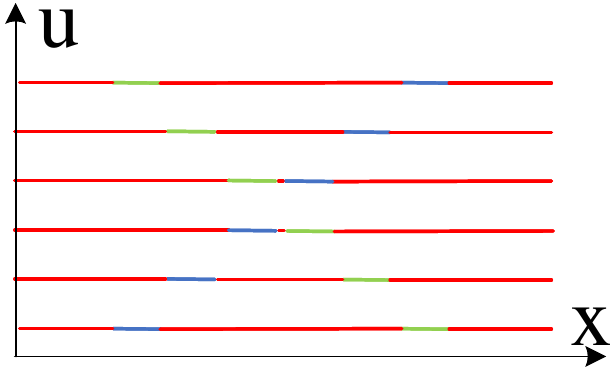}
}
\subfigure[Allowed occ.]{
	\label{fig:wrong_occ:b}
	\includegraphics[width=25mm]{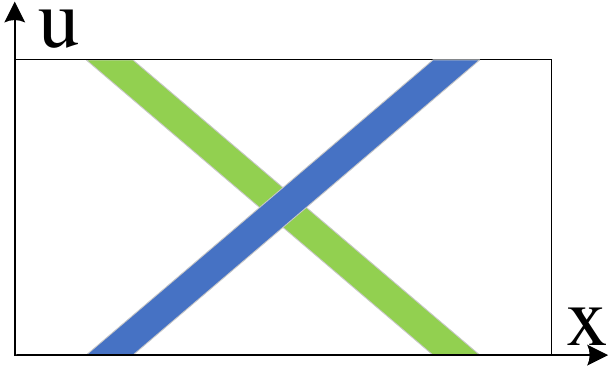}
}
\subfigure[Forbidden occ.]{
	\label{fig:wrong_occ:c}
	\includegraphics[width=25mm]{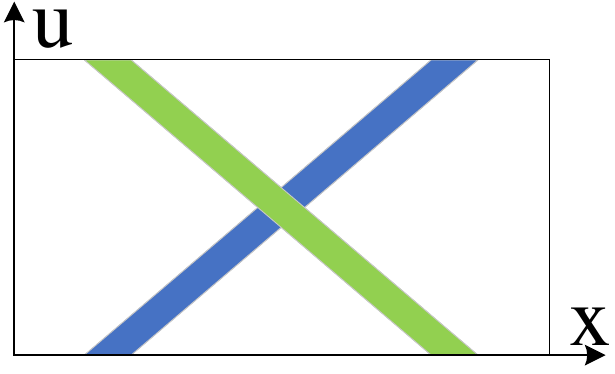}
}
\end{center}
\caption{Flip operation causing the network to learn incorrect occlusions. (a) Recorded light field\cite{levin2008understanding}: The red lines refer to sampled views while the blue/green lines account for the EPI lines; (b) Occlusion after reconstruction; (c) Incorrect occlusion caused by the flip operation.}
\label{fig:wrong_occlusion}
\end{figure}

Note that, the flip operation as commonly used for data augmentation in traditional image super-resolution\cite{dong2016image}, can not be applied in EPI super-resolution. This is since, in its traditional form, the network will learn a wrong occlusion. Note that, as shown in Fig.\ref{fig:wrong_occ:a}, the intersection between foreground and background is lost after the light field sampling. Thus, the flip operation will lead the wrong occlusion to be learnt, causing forbidden occlusions to appear in the reconstructed light fields. This is shown in Fig.\ref{fig:wrong_occ:c}, where an incorrect light field is reconstructed when the foreground is occluded by the background.

\section{Experimental Results}
\label{sec:experiments}
We compare our method with a combination of state-of-the-art light field reconstruction methods such as EPICNN\cite{wu2017light} and LFST\cite{vagharshakyan2018light} and deep-learning based image super-resolution methods such as SRCNN\cite{dong2016image}. All the results shown here are evaluated using the code released by the authors. 

We evaluate the performance of the proposed method both on synthetic and real light fields. All the quantitative comparisons shown here are average values for all the views under study. As our network is trained on synthetic data, to be fair, the data introduced in Sec.\ref{sec:experiment:syn_data} is only used to validate the efficacy of the LSTM layers. Real-world data from camera array \cite{stanford_lf_web,kim2013scene} is used for comparison. Here we have not used the light fields from the Lytro Illum camera due to their small disparity.

\subsection{Synthetic data}

\begin{figure}[h]
\begin{center}
\centering
\includegraphics[width=52mm]{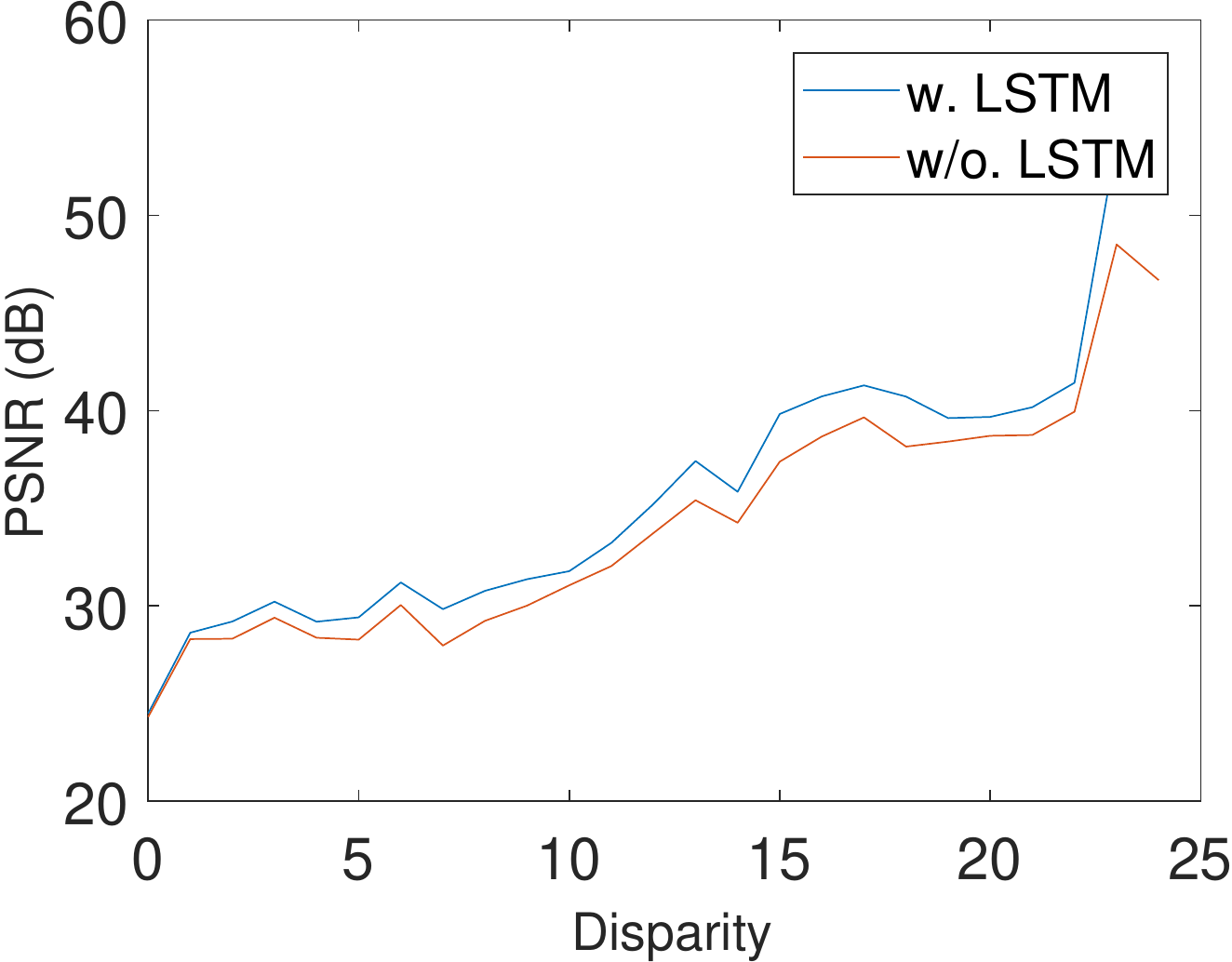}
\end{center}
\caption{Performance of our network with and without LSTM layers as a function of disparity.}
\label{fig:lstm_wolstm_difdis_dingliang}
\end{figure}

In Tab.\ref{tab:lstm_eff_cmp}, we show the quantitative comparison between the results yielded by our network with and without the LSTM layers. Fig.\ref{fig:lstm_wolstm_difdis_dingliang} shows a plot of the PSNR for both settings as a function of disparity. Note that, as expected, our network with LSTM layers outperforms the one without over almost all of the disparity range. 

\begin{figure}[h]
\begin{center}
\centering
\includegraphics[width=75mm]{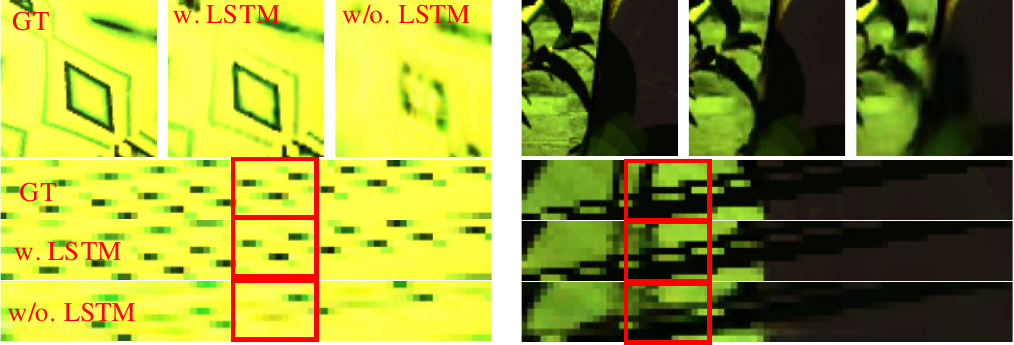}
\end{center}
\caption{Qualitative comparisons between the ground truth and the results from networks with and without LSTM layers, respectively. Compared with the one without LSTM, LSTM provides more clear novel views and more accurate EPI lines.}
\label{fig:lstm_wolstm_dingxing}
\end{figure}

Furthermore, Fig.\ref{fig:lstm_wolstm_dingxing} shows qualitative results. Notice that the network with LSTM delivers more detail than the one without LSTM in the reconstructed views. This is mainly due to the fact that discrete EPI lines in these areas are over-smoothed as shown in the bottom EPI comparison in Fig.\ref{fig:lstm_wolstm_dingxing} when LSTM is not included. This is consistent with the notion that LSTM can better cope with the ``discontinuous continuity'' in EPIs.  

\begin{table}[h]
\small
\begin{center}
\caption{Quantitative comparison between the results yielded by our network with and without LSTM layers.}
\label{tab:lstm_eff_cmp}
\begin{tabular}{|l|c|c|}
\hline
	& PSNR (dB) & SSIM\\
\hline
w. LSTM & \textbf{28.34} & \textbf{0.886}  \\
\hline
w/o. LSTM &27.75 & 0.863\\
\hline
\end{tabular}
\end{center}
\end{table}

\subsection{Real data}
\subsubsection{Comparison with State-of-the-arts}
For purposes of  comparison on real-data, we have used the Stanford light field dataset (SLFD)\cite{stanford_lf_web}. 
In order to compare the performance of different methods more fairly, we zoom out each view of the SLFD to $0.2,\:0.3,\:0.4,\:0.5$ of the original size. Recall that the disparity range decreases with respect to  the zoom out factor. Thus, for the readers reference, we show the disparity ranges of the originally sized view are shown in Tab. \ref{tab:details:cameraarray}.

\begin{table}[t]
\scriptsize
\begin{center}
\caption{Disparity ranges of the SLFD \cite{stanford_lf_web}.}
\label{tab:details:cameraarray}
\begin{tabular}{|l|c|c|c|c|c|c|}
\hline
Data &Amethyst & Bulldozer & Bunny &Chess &Lego & Truck\\
\hline
Dis. &$[-6,5]$ &$[-4,20]$ & $[-8,5]$  &$[0,7]$ &$[-9,7]$ &$[0,3]$ \\
\hline
\end{tabular}
\end{center}
\end{table}

\begin{table}[tbp]
\scriptsize
\begin{center}
\caption{Quantitative comparison on SLFD.}
\label{tab:real:psnr_ssim}
\begin{tabular}{|l|c|c|c|c|c|c|}
\hline
\multirow{2}{*}{Factor} &\multicolumn{2}{c|}{EPICNN\cite{wu2017light}+SRCNN\cite{dong2016image}} & \multicolumn{2}{c|}{LFST\cite{vagharshakyan2018light}+SRCNN\cite{dong2016image}} & \multicolumn{2}{c|}{Ours}\\
\cline{2-7}
		& PSNR(dB) & SSIM & PSNR & SSIM & PSNR & SSIM \\
\hline
0.2 & 28.77 & 0.898 & \textbf{32.38} & 0.945 &32.17 &\textbf{0.956}\\
0.3 & 28.28 & 0.891 & 32.20 & 0.940 &\textbf{32.85} &\textbf{0.958}\\
0.4 & 27.97 & 0.889 & 32.27 & 0.944 &\textbf{33.44} &\textbf{0.961}\\
0.5 & 27.71 & 0.888 & 31.19 & 0.935 &\textbf{33.70} &\textbf{0.960}\\
\hline
\end{tabular}
\end{center}
\end{table}


\begin{figure*}[!]
\begin{center}
\centering
\subfigure[Bulldozer]{
	\label{fig:real:stanford:Bulldozer}
	\includegraphics[width=158mm]{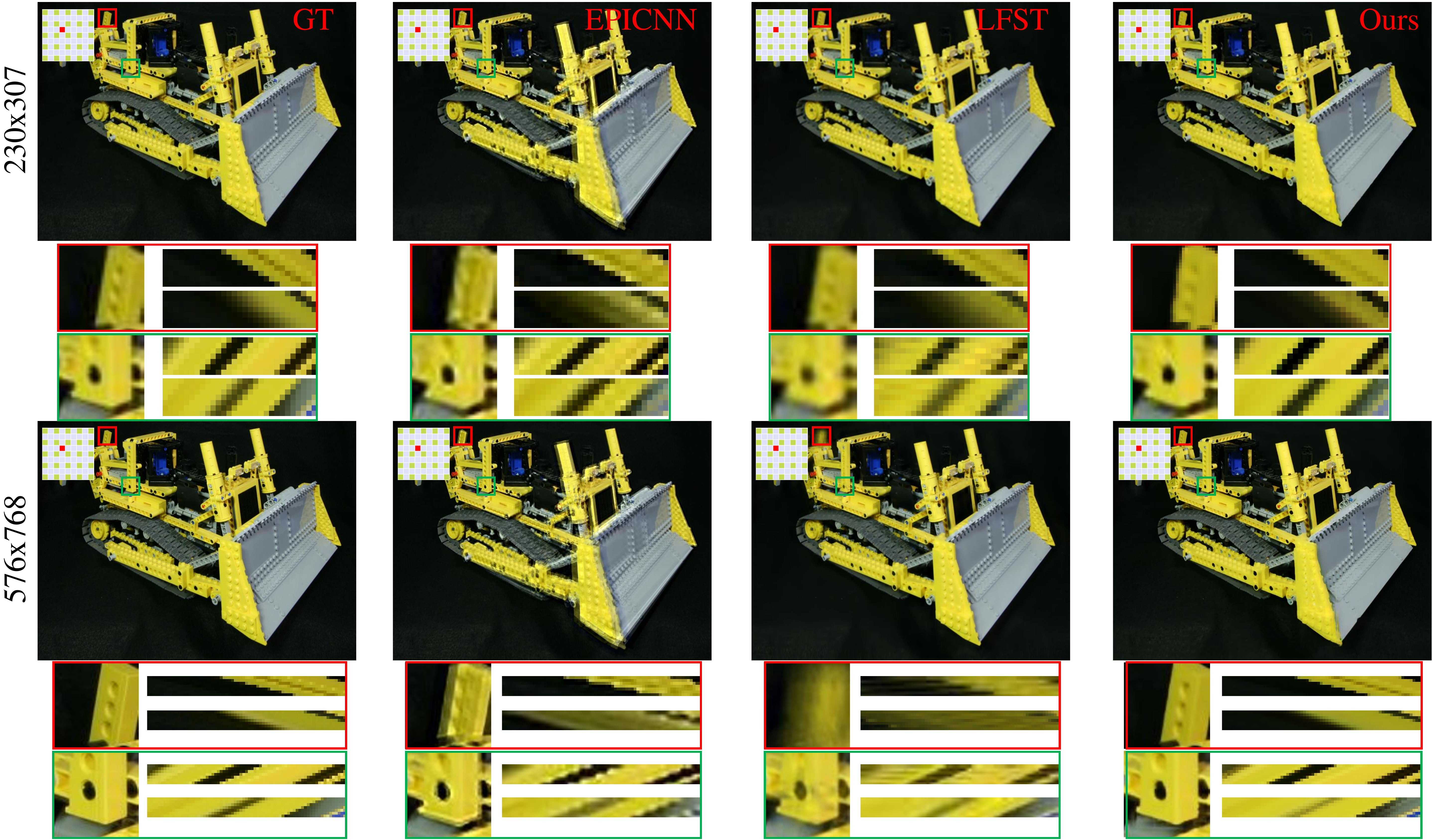}
}\\
\subfigure[Legoknights]{
	\label{fig:real:stanford:Legoknights}
	\includegraphics[width=158mm]{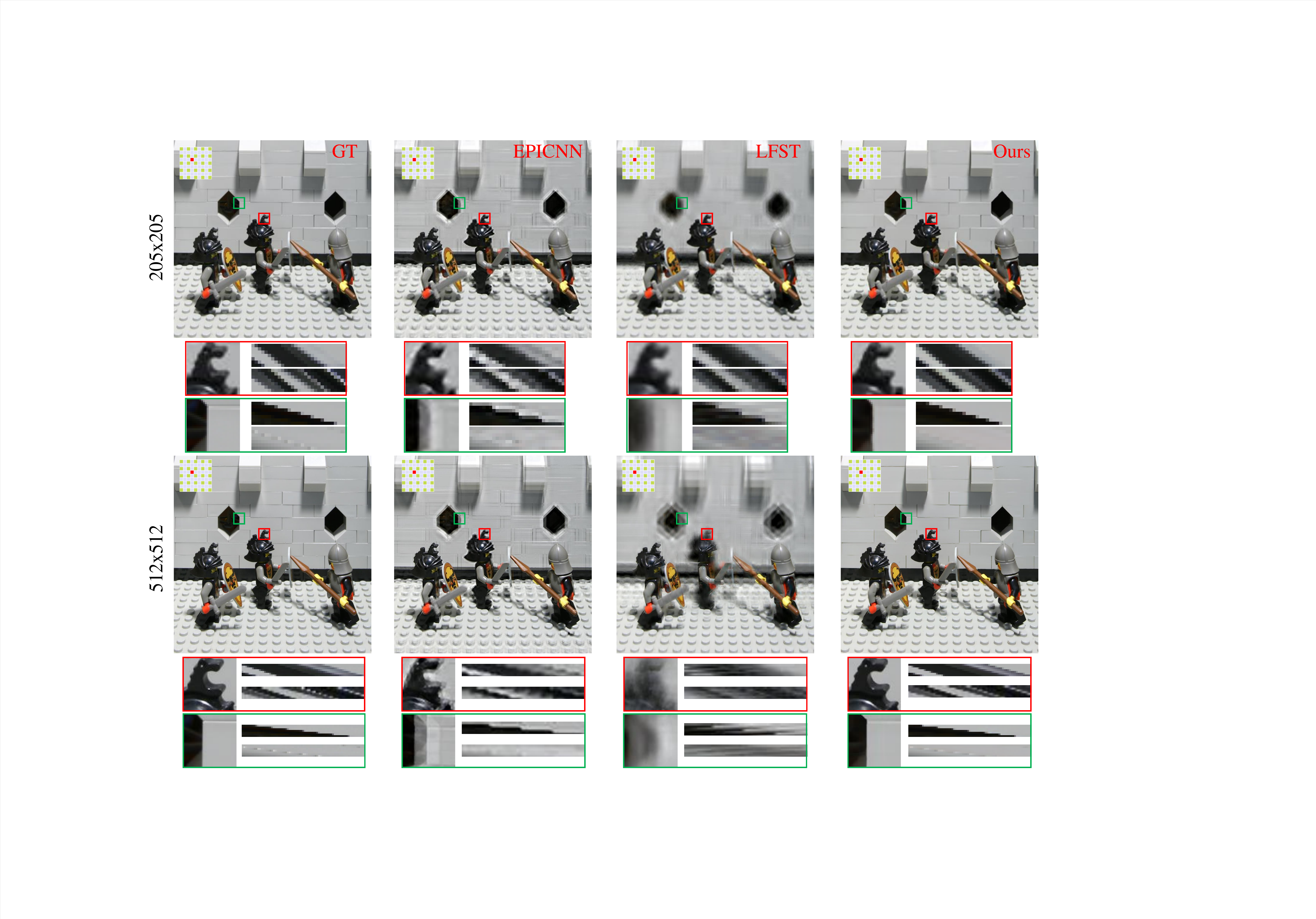}
}
\end{center}
\caption{Qualitative comparison on the Bulldozer and the Lego. For each light field, the first and second rows show the results at different resolution inputs. For each of the zoom in areas in red and green, the left panel shows the reconstructed view, while the right two rectangular panels shwo the horizontal and vertical EPIs computed from the reconstructed light field.}
\label{fig:real:stanford_12}
\end{figure*}

Tab. \ref{tab:real:psnr_ssim} shows a quantitative comparisons of our method with respect to the alternatives on SLFD. Note that our method outperforms the alternatives at almost all of the zoom out factors. Although our network only employs synthetic data during the training process, it  shows a good generalization ability as applied to unseen camera array data. 
Fig.\ref{fig:real:stanford_12} shows qualitative demonstrations\footnote{More results are provided in the supplementary material.}. 
For each scene in Fig.\ref{fig:real:stanford_12}, the first and second rows refer to results for zoom out factors $0.2$ and $0.5$, respectively. Note that EPICNN and LFST achieve in general similar performance as ours at small zoom out factors. However, their performance decreases at large zoom out factors and they tend to over-smooth object boundaries. Ours, in the other hand, can always maintain sharp object boundaries at both small and large zoom out factors. For example, in Fig.\ref{fig:real:stanford:Bulldozer}, the boundaries of Bulldozer are all preserved well. However,  previous methods often fail at large zoom out factors. 

\noindent \textbf{Ours vs EPICNN:} Compared with the state-of-the-art, our method has achieves at least a 3dB lead (32.17 vs 28.77). This advantage increases with the disparity. Fig.\ref{fig:real:stanford:Bulldozer} gives a better comparison in larger disparity areas. There are serious ghosting in the shovel boundaries recovered by EPICNN. Since the maximum disparity at these areas is about $20$ pixels, the EPI consistency on the shovel is lost by EPICNN, while our result remains sharp.

\noindent \textbf{Ours vs LFST:} Despite LFST achieves good results in large positive disparity regions (such as the shovel boundaries in Fig.\ref{fig:real:stanford:Bulldozer}), the results at negative disparity regions are somewhat mediocre. The best example is Fig.\ref{fig:real:stanford:Legoknights}, where the areas in front of and behind the toy warriors have positive and negative disparities, respectively. LFST induces ghosting in large negative disparity areas. In contrast, our method produces consistent results in both positive and negative areas. This is thanks to the shearing (Eqn.\ref{eqn:data_augmentation:shearing}) data augmentation and the LSTM's ability to model EPIs. Furthermore, in contrast with our approach, LFST often generates artifacts in texture boundaries, as shown in some of the green boxes on Fig. \ref{fig:real:stanford:Bulldozer}.

\subsubsection{Disparity vs Resolution}
As shown in Tab. \ref{tab:real:psnr_ssim}, the performance of our method increases in a manner commensurate with the scale. To better illustrate this, we conducted another two experiments by fixing the spatial resolution and disparity range, respectively. To this end, we have used the Disney light field dataset\cite{kim2013scene}\footnote{Details of the light fields are provided in the supplementary material.}. Since it has high angular and spatial resolutions, we have conducted experiments by controlling the number of skipped views. Generally, a larger number of skipped views leads to a larger disparity range and lower resolution.

\begin{figure}[tbp]
\begin{center}
\centering
	\includegraphics[width=75mm]{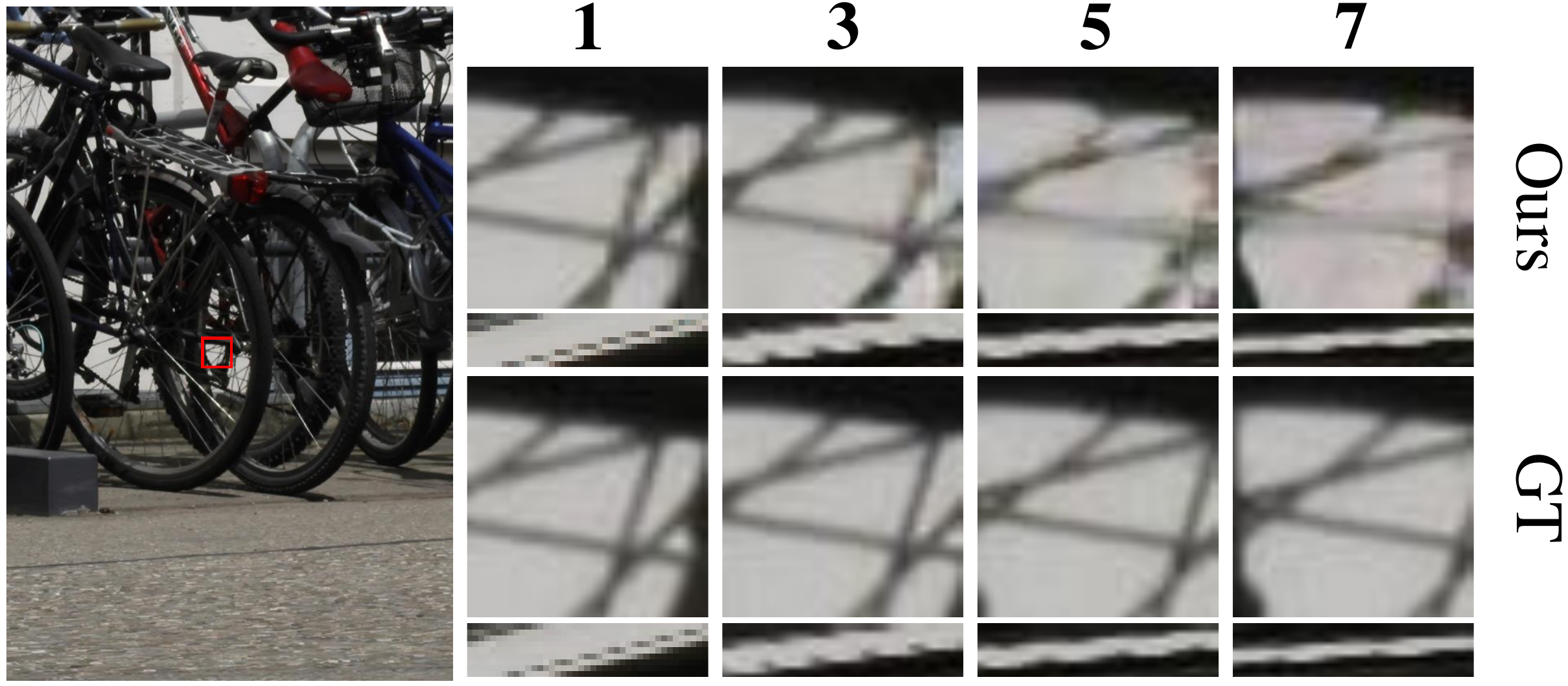}
\end{center}
\caption{Results yielded by our method with fixed resolution for several disparity values.}
\label{fig:real:disney:changedis_fixres}
\end{figure}

Tab. \ref{tab:psnr_ssim:disney:fixres_changedis} and \ref{tab:psnr_ssim:disney:fixdis_changeres} show quantitative comparisons of the proposed method in these two experiments, respectively. The performance of our method decreases with increases in disparity. The performance increases with increments in resolution. In large disparity areas with complex textures, the EPI consistency is very weak, so a small error in EPI reconstruction leads to heavy artifacts in the reconstructed view. This can be seen in the reconstructed EPIs which are very close to the ground truth both in small disparity and large disparity cases in Fig.\ref{fig:real:disney:changedis_fixres}. However, the corresponding view is well recovered at small disparity being overly smoothed at large disparity. Regardless, both the angular and spatial resolution are improved by our network. Further, the defects induced in the angular domain are compensated by the super-resolution in the spatial domain. As a result, the performance of our method seemingly increases as the zoom out factor increases in Tab. \ref{tab:real:psnr_ssim}.

\begin{table}[tbp]
\footnotesize
\begin{center}
\caption{Comparison of our method with the alternatives with fixed resolution and increasing disparity.}
\label{tab:psnr_ssim:disney:fixres_changedis}
\begin{tabular}{|l|c|c|c|c|}
\hline
\multirow{2}{*}{Skipped views} &\multicolumn{2}{c|}{Bikes} & \multicolumn{2}{c|}{Couch} \\
\cline{2-5}
		& PSNR(dB) & SSIM & PSNR & SSIM \\
\hline
1 (low dis.) & $35.35$ & $0.974$ & $35.59$ & $0.948$\\
3 & $32.06$ & $0.947$ & $34.23$ & $0.931$\\
5 & $29.28$ & $0.909$ & $31.52$ & $0.904$\\
7 (high dis.)& $28.16$ & $0.893$ & $29.69$ & $0.898$\\
\hline
\end{tabular}
\end{center}
\end{table}
\begin{table}[tbp]
\footnotesize
\begin{center}
\caption{Comparisons of our method with the alternatives with fixed disparity and increasing resolution.}
\label{tab:psnr_ssim:disney:fixdis_changeres}
\begin{tabular}{|l|c|c|c|c|}
\hline
\multirow{2}{*}{Skipped views} &\multicolumn{2}{c|}{Bikes} & \multicolumn{2}{c|}{Couch} \\
\cline{2-5}
		& PSNR(dB) & SSIM & PSNR & SSIM \\
\hline
7  (low res.)& $29.68$ & $0.923$ & $33.06$ & $0.872$\\
5  & $30.87$ & $0.942$ & $33.63$ & $0.886$\\
3  & $32.87$ & $0.963$ & $34.17$ & $0.906$\\
1  (high res.)& $35.34$ & $0.975$ & $35.57$ & $0.948$\\
\hline
\end{tabular}
\end{center}
\end{table}

\section{Conclusion}
In this paper we have shown that,  since most 3D points in a scene are generally defocused in an LFC, contrary to a common belief {\em as per} the spatial-angular resolution trade-off, the resolution of an LFC is in fact larger than its number of micro-lenses. This new insight provides a theoretical basis to overcome the barrier of ``spatio-angular trade-off''. By analyzing the light path in an LFC, we have identified two different types of ``continuity'' in EPIs.  We have proposed a novel CNN-LSTM network to practically super-resolve a high resolution light field in both, the spatial and angular axes.  Experiments on synthetic and real-world light fields validate the superiority of the proposed method in large disparity areas, outperforming most of the state-of-the-art methods.  

{\small
\bibliographystyle{ieee}
\bibliography{egbib}

\begin{thebibliography}{10}\itemsep=-1pt

\bibitem{stanford_lf_web}
The new stanford light field archive.
\newblock \url{http://lightfield.stanford.edu/lfs.html}.

\bibitem{bishop2012light}
T.~E. Bishop and P.~Favaro.
\newblock The light field camera: Extended depth of field, aliasing, and
  superresolution.
\newblock {\em TPAMI}, 34(5):972--986, 2012.

\bibitem{broxton2013wave}
M.~Broxton, L.~Grosenick, S.~Yang, N.~Cohen, A.~Andalman, K.~Deisseroth, and
  M.~Levoy.
\newblock Wave optics theory and 3-d deconvolution for the light field
  microscope.
\newblock {\em Optics express}, 21(21):25418--25439, 2013.

\bibitem{chai2000plenoptic}
J.-X. Chai, X.~Tong, S.-C. Chan, and H.-Y. Shum.
\newblock Plenoptic sampling.
\newblock In {\em SIGGRAPH}, pages 307--318. ACM Press/Addison-Wesley
  Publishing Co., 2000.

\bibitem{chaurasia2013depth}
G.~Chaurasia, S.~Duchene, O.~Sorkine-Hornung, and G.~Drettakis.
\newblock Depth synthesis and local warps for plausible image-based navigation.
\newblock {\em TOG}, 32(3):30, 2013.

\bibitem{chaurasia2011silhouette}
G.~Chaurasia, O.~Sorkine, and G.~Drettakis.
\newblock Silhouette-aware warping for image-based rendering.
\newblock In {\em CGF}, volume~30, pages 1223--1232. Wiley Online Library,
  2011.

\bibitem{dong2016image}
C.~Dong, C.~C. Loy, K.~He, and X.~Tang.
\newblock Image super-resolution using deep convolutional networks.
\newblock {\em TPAMI}, 38(2):295--307, 2016.

\bibitem{eisemann2008floating}
M.~Eisemann, B.~De~Decker, M.~Magnor, P.~Bekaert, E.~De~Aguiar, N.~Ahmed,
  C.~Theobalt, and A.~Sellent.
\newblock Floating textures.
\newblock In {\em CGF}, volume~27, pages 409--418. Wiley Online Library, 2008.

\bibitem{georgiev2006spatio}
T.~Georgiev, K.~C. Zheng, B.~Curless, D.~Salesin, S.~K. Nayar, and C.~Intwala.
\newblock Spatio-angular resolution tradeoffs in integral photography.
\newblock {\em Rendering Techniques}, 2006(263-272):21, 2006.

\bibitem{georgiev2009superresolution}
T.~G. Georgiev and A.~Lumsdaine.
\newblock Superresolution with plenoptic 2.0 cameras.
\newblock In {\em Signal recovery and synthesis}, page STuA6. Optical Society
  of America, 2009.

\bibitem{goesele2010ambient}
M.~Goesele, J.~Ackermann, S.~Fuhrmann, C.~Haubold, R.~Klowsky, D.~Steedly, and
  R.~Szeliski.
\newblock Ambient point clouds for view interpolation.
\newblock In {\em TOG}, volume~29, page~95. ACM, 2010.

\bibitem{gortler1996lumigraph}
S.~J. Gortler, R.~Grzeszczuk, R.~Szeliski, and M.~F. Cohen.
\newblock The lumigraph.
\newblock In {\em SIGGRAPH}, pages 43--54. ACM, 1996.

\bibitem{heber2016u}
S.~Heber, W.~Yu, and T.~Pock.
\newblock U-shaped networks for shape from light field.
\newblock In {\em BMVC}, volume~3, page~5, 2016.

\bibitem{heber2017neural}
S.~Heber, W.~Yu, and T.~Pock.
\newblock Neural epi-volume networks for shape from light field.
\newblock In {\em IEEE CVPR}, pages 2252--2260, 2017.

\bibitem{hochreiter1997long}
S.~Hochreiter and J.~Schmidhuber.
\newblock Long short-term memory.
\newblock {\em Neural computation}, 9(8):1735--1780, 1997.

\bibitem{LearningViewSynthesis}
N.~K. Kalantari, T.-C. Wang, and R.~Ramamoorthi.
\newblock Learning-based view synthesis for light field cameras.
\newblock {\em TOG}, 35(6), 2016.

\bibitem{kim2013scene}
C.~Kim, H.~Zimmer, Y.~Pritch, A.~Sorkine-Hornung, and M.~H. Gross.
\newblock Scene reconstruction from high spatio-angular resolution light
  fields.
\newblock {\em TOG}, 32(4):73--1, 2013.

\bibitem{kim2016accurate}
J.~Kim, J.~Kwon~Lee, and K.~Mu~Lee.
\newblock Accurate image super-resolution using very deep convolutional
  networks.
\newblock In {\em IEEE CVPR}, pages 1646--1654, 2016.

\bibitem{krizhevsky2012imagenet}
A.~Krizhevsky, I.~Sutskever, and G.~E. Hinton.
\newblock Imagenet classification with deep convolutional neural networks.
\newblock In {\em NIPS}, pages 1097--1105, 2012.

\bibitem{lai2017deep}
W.-S. Lai, J.-B. Huang, N.~Ahuja, and M.-H. Yang.
\newblock Deep laplacian pyramid networks for fast and accurate
  super-resolution.
\newblock In {\em IEEE CVPR}, pages 624--632, 2017.

\bibitem{lecun1990handwritten}
Y.~LeCun, B.~E. Boser, J.~S. Denker, D.~Henderson, R.~E. Howard, W.~E. Hubbard,
  and L.~D. Jackel.
\newblock Handwritten digit recognition with a back-propagation network.
\newblock In {\em NIPS}, pages 396--404, 1990.

\bibitem{levin2010linear}
A.~Levin and F.~Durand.
\newblock Linear view synthesis using a dimensionality gap light field prior.
\newblock In {\em IEEE CVPR}, pages 1831--1838, 2010.

\bibitem{levin2008understanding}
A.~Levin, W.~T. Freeman, and F.~Durand.
\newblock Understanding camera trade-offs through a bayesian analysis of light
  field projections.
\newblock In {\em Springer ECCV}, pages 88--101, 2008.

\bibitem{levoy1996light}
M.~Levoy and P.~Hanrahan.
\newblock Light field rendering.
\newblock In {\em SIGGRAPH}, pages 31--42. ACM, 1996.

\bibitem{lin2004geometric}
Z.~Lin and H.-Y. Shum.
\newblock A geometric analysis of light field rendering.
\newblock {\em IJCV}, 58(2):121--138, 2004.

\bibitem{lytro_web}
Lytro.
\newblock Lytro redefines photography with light field cameras.
\newblock \url{http://www.lytro.com}, 2011.

\bibitem{nair2010rectified}
V.~Nair and G.~E. Hinton.
\newblock Rectified linear units improve restricted boltzmann machines.
\newblock In {\em ICML}, pages 807--814, 2010.

\bibitem{ng2006digital}
R.~Ng et~al.
\newblock {\em Digital light field photography}.
\newblock stanford university Stanford, CA, 2006.

\bibitem{penner2017soft}
E.~Penner and L.~Zhang.
\newblock Soft 3d reconstruction for view synthesis.
\newblock {\em TOG}, 36(6):235, 2017.

\bibitem{povray_web}
POV-ray.
\newblock \url{http://www.povray.org/}.

\bibitem{raytrix_web}
Raytrix.
\newblock $\infty$ raytrix.
\newblock \url{http://www.raytrix.de}, 2012.

\bibitem{shi2014light}
L.~Shi, H.~Hassanieh, A.~Davis, D.~Katabi, and F.~Durand.
\newblock Light field reconstruction using sparsity in the continuous fourier
  domain.
\newblock {\em TOG}, 34(1):12, 2014.

\bibitem{srinivasan2017learning}
P.~P. Srinivasan, T.~Wang, A.~Sreelal, R.~Ramamoorthi, and R.~Ng.
\newblock Learning to synthesize a 4d rgbd light field from a single image.
\newblock In {\em IEEE ICCV}, volume~2, page~6, 2017.

\bibitem{povray_resource}
G.~Tran.
\newblock Oyonale - 3d art and graphic experiments.
\newblock \url{http://www.oyonale.com/}.

\bibitem{vagharshakyan2018light}
S.~Vagharshakyan, R.~Bregovic, and A.~Gotchev.
\newblock Light field reconstruction using shearlet transform.
\newblock {\em TPAMI}, 40(1):133--147, 2018.

\bibitem{wang2017light}
T.-C. Wang, J.-Y. Zhu, N.~K. Kalantari, A.~A. Efros, and R.~Ramamoorthi.
\newblock Light field video capture using a learning-based hybrid imaging
  system.
\newblock {\em TOG}, 36(4):133, 2017.

\bibitem{wang2018end}
Y.~Wang, F.~Liu, Z.~Wang, G.~Hou, Z.~Sun, and T.~Tan.
\newblock End-to-end view synthesis for light field imaging with pseudo 4dcnn.
\newblock In {\em Springer ECCV}, pages 340--355, 2018.

\bibitem{wanner2014variational}
S.~Wanner and B.~Goldluecke.
\newblock Variational light field analysis for disparity estimation and
  super-resolution.
\newblock {\em TPAMI}, 36(3):606--619, 2014.

\bibitem{wu2017light}
G.~Wu, M.~Zhao, L.~Wang, Q.~Dai, T.~Chai, and Y.~Liu.
\newblock Light field reconstruction using deep convolutional network on epi.
\newblock In {\em IEEE CVPR}, volume 2017, page~2, 2017.

\bibitem{xingjian2015convolutional}
S.~Xingjian, Z.~Chen, H.~Wang, D.-Y. Yeung, W.-K. Wong, and W.-c. Woo.
\newblock Convolutional lstm network: A machine learning approach for
  precipitation nowcasting.
\newblock In {\em NIPS}, pages 802--810, 2015.

\bibitem{yeung2018fast}
H.~W.~F. Yeung, J.~Hou, J.~Chen, Y.~Y. Chung, and X.~Chen.
\newblock Fast light field reconstruction with deep coarse-to-fine modelling of
  spatial-angular clues.
\newblock In {\em Springer ECCV}, pages 137--152, 2018.

\bibitem{yoon2015learning}
Y.~Yoon, H.-G. Jeon, D.~Yoo, J.-Y. Lee, and I.~So~Kweon.
\newblock Learning a deep convolutional network for light-field image
  super-resolution.
\newblock In {\em IEEE ICCV Workshops}, pages 24--32, 2015.

\bibitem{zhou2018stereo}
T.~Zhou, R.~Tucker, J.~Flynn, G.~Fyffe, and N.~Snavely.
\newblock Stereo magnification: Learning view synthesis using multiplane
  images.
\newblock {\em arXiv preprint arXiv:1805.09817}, 2018.

\end{thebibliography}
}

\end{document}